# Performance Analysis of Keypoint Detectors and Binary Descriptors Under Varying Degrees of Photometric and Geometric Transformations

Shuvo Kumar Paul, Pourya Hoseini, Mircea Nicolescu, Monica Nicolescu
University of Nevada, Reno, USA
Email: {shuvo.k.paul, hoseini}@nevada.unr.edu, {mircea, monica}@cse.unr.edu

*Abstract*— Detecting image correspondences by feature matching forms the basis of numerous computer vision applications. Several detectors and descriptors have been presented in the past, addressing the efficient generation of features from interest points (keypoints) in an image. In this paper, we investigate eight binary descriptors (AKAZE, BoostDesc, BRIEF, BRISK, FREAK, LATCH, LUCID, and ORB) and eight interest point detector (AGAST, AKAZE, BRISK, FAST, HarrisLapalce, KAZE, ORB, and StarDetector). We have decoupled the detection and description phase to analyze the interest point detectors and then evaluate the performance of the pairwise combination of different detectors and descriptors. We conducted experiments on a standard dataset and analyzed the comparative performance of each method under different image transformations. We observed that: (1) the FAST, AGAST, ORB detectors were faster and detected more keypoints, (2) the AKAZE and KAZE detectors performed better under photometric changes while ORB was more robust against geometric changes, (3) in general, descriptors performed better when paired with the KAZE and AKAZE detectors, (4) the BRIEF,LUCID, ORB descriptors were relatively faster, and (5) none of the descriptors did particularly well under geometric transformations, only BRISK, FREAK, and AKAZE showed reasonable resiliency.

*Index Terms*— interest point detector, keypoint detector, descriptor, evaluation, image features, binary descriptor

## I. Introduction

Detecting interest points and locating correspondences between two images play a crucial role in numerous computer vision applications such as: object detection and pose estimation [1], visual odometry [2], simultaneous localization and mapping [3], augmented reality [4], image mosaicing and panorama stitching [5]. Generally, interest point detectors are used to extract the candidate points, and descriptors are used to form the description of the image structure surrounding that keypoint. Subsequently, a comparison is done on the extracted descriptors in different images using relevant similarity metrics to determine the corresponding interest points. This is based on the assumption that the same feature will be detected in two or more different images of the same scene, and that descriptors extract the essential information that encodes the visual information in the regions surrounding the interest points. Ideally, this description of the same scene for points seen from different viewpoints or subject to other image transformations should be similar. Therefore, an ideal detector should detect the same keypoints present in the scene and the descriptor should be invariant to any photometric or geometric image transformation.

In 2004, D. Lowe published his seminal work SIFT [6], and later H. Bay et al. presented SURF [7], a speeded-up version of SIFT. Both of these methods are two of the most widely used techniques for local feature extraction. These methods helped address many computer vision problems with great accuracy and are known for their robustness under image transformations. However, this reliability of performance comes at a high computational cost as storing high dimensional descriptors in floating point representation not only takes a significant amount of memory but also requires more time for matching descriptors. Although a few other variants of SIFT and SURF were proposed to address these limitations, they were not efficient enough for devices with limited computational resources.

In recent times, the wide adoption of handheld devices, e.g. mobile phones, digital cameras, etc. has popularized vision-based applications that include visual search, augmented reality, image filters, wearable computing, etc. However, most of these devices do not have the ample computational capability required to maintain a minimum execution speed for such applications, due to the manipulation of high dimensional floating point numbers. This warranted for more efficient techniques for interest point detection and feature description.

During the last two decades, several new feature detectors and binary descriptors have been developed. All of them had a significant improvement in terms of execution speed but are not as robust compared to SIFT and SURF. Although some of these feature detectors and descriptors work well under certain image transformations, there is no one solution that is feasible in all situations. Moreover, different devices have various computing resources and different applications have specific requirements, thus selecting a feature-detector-descriptor

---

The code is available at https://github.com/paul-shuvo/detector-descriptor-evaluation

combination that balances speed and accuracy is a critical decision for developing a solution.

For this evaluation, we selected eight detectors and eight binary descriptors that are included in the popular OpenCV library. The detectors are HarrisLaplace [8], FAST [9], AGAST [10], KAZE [11], AKAZE [12], ORB [13], BRISK [14], and StarDetector [15], and the binary descriptors are BRIEF [16], AKAZE [12], BoostDesc [17,18], LATCH [19], LUCID [20], BRISK [14], ORB [13], and FREAK [21].

We evaluated the detectors in terms of processing time, the number of keypoints detected, and repeatability ratio with respect to different degrees of photometric changes: blur, illumination, jpeg-compression. All of these detectors were then combined with all the descriptors (except for AKAZE which only supports KAZE and AKAZE detectors) to measure the combined performance under different photometric and geometric image transformations.

Our goal in this work is to present an analysis of the recent developments in feature detectors and descriptors that can guide on selecting the optimum detector-descriptor combination that requires lower computing resources. Additionally, we provide a publicly available evaluation framework that can not only be used to compare and analyze other detector and descriptor techniques but also to evaluate the already included methods on different image sequences.

This paper is outlined as follows: in the next section, we provide a brief overview of previous evaluations. Next, we describe the dataset and image sequences. The following chapters include our evaluation containing experimental results and observations. Finally, we conclude this paper by summarizing our work.

## II. RELATED WORK

In [22], Christensen et al. presented empirical evaluation techniques that allow objective comparative assessment of computer vision algorithms. Schmid et al. [23] evaluated interest point detectors based on repeatability rate criterion and information content criterion; repeatability rate is defined as the number of points repeated between two images with respect to the total number of detected points and information content is a measure of the distinctiveness of an interest point. Mikolajczyk et al. [24] provided a set of benchmark image sequences (the Oxford dataset) for testing the effects of blur, compression, exposure, scale/rotation, and perspective change; since these image sequences have been widely adopted in vision research, we used them in our evaluation for compatibility.

In 2005, Mikolajczyk and Schmid [25] compared several descriptors computed for local interest regions and investigated if and how their performance is depended on the interest region detectors. Later, Misksik and Mikolajczyk evaluated mostly binary descriptors in [26], together with other detectors to determine detector-descriptor dependencies. Heinly et al. [27] compared the performance of three binary descriptors with other gradient based descriptors by pairing them all with different interest point detectors.

Other comparison articles were dedicated to a specified task. Moreels and Perona [28] explored the performance of several feature detectors and descriptors in matching 3D object features across different viewpoints and lighting conditions. Fraundorfer et al. [29] presented a method to evaluate the performance of local detectors that allows the automatic verification of detected corresponding points and regions for non-planar scenes. Gauglitz et al. [30] compared both interest point detectors and feature descriptors in isolation along with all the detector-descriptor combinations for visual tracking. Ali et al. [31] analyzed the performance of various feature detectors and descriptors for panorama image stitching. Tareen et al. [32] did a comparative study of popular detector-descriptor methods for image registration. Shi et al. [33] investigated several feature descriptors under rotational changes in various image features extracted from monocular thermal and photo camera images. Dahl et al. [34] evaluated some of the most popular descriptor and detector combinations on the DTU Robot dataset, which is a systematic data aimed at two view matching. Mandalin [35] compared six descriptors applied to face recognition using different distance measures with Nearest Neighbor and SVM as classifiers. Aanæs et al. [36] evaluated detectors based on spatial invariance of interest points under changing acquisition parameters of a large dataset.

SIFT and SURF are well studied and often used as a benchmark in the literature. It is well established that both methods generally perform well under different image transformations but have slower run time. Rather than comparing with these two methods or their variants, we focus on doing a comparative analysis on the faster detector and binary descriptor methods.

## III. DATASET

Our evaluation was done on the Oxford affine dataset [25] which contains eight sequences, each corresponding to different photometric and geometric image transformations that include change of blur, viewpoint, rotation and zoom, jpeg-compression, and illumination. Each image has a medium resolution (approximately 800 x 640 pixels). Each image sequence is composed of six images (enumerated from 0 to 5) and the ground truth homographies between the first and the rest of the images in the sequence. For simplicity we would refer the first image as Train image **T** and the rest of the images in the sequence as Query image **Q**. Each subsequent query image has more variance than the prior image in the sequence, for example, $Q_{i+1}$ would contain more variance than $Q_i$ ($Q_0$, in this case, would be image **T**). Although the image numbers on the x-axis in the Fig. 1, and Fig. 2 appear in linear scale, this does not represent linear change – it simply indicates the addition of more variance on the subsequent images in the sequence.

## IV. EVALUATION

We measured the number of detected keypoints, and repeatability rate of each detector under different degrees of photometric changes along with the average processing time. Subsequently, we paired the descriptors with all the other detectors (except for AKAZE which only supports KAZE and AKAZE detector) and measured their average processing time and average accuracy, precision, and recall. We have selected the default variant for each method implemented in OpenCV; Table I arranges the memory size of the default variants of descriptors that are evaluated in ascending order. We have used the following abbreviations to accommodate the results on Table IV and V: HL (HarrisLaplace), FT(FAST), AG (AGAST), KZ (KAZE), AZ (AKAZE), OB(ORB), BK (BRISK), SD (StarDetector), BF (BRIEF), BD (BoostDesc), LT (LATCH), LD (LUCID), BK (BRISK), OB (ORB), and FK (FREAK). Color gradients are used on Table III, IV, and V to illustrate the comparative performance of the methods arranged in rows; darker value indicates better performance.

each stage of added variance. Fig. 1(a, b, c) illustrates the change in number of keypoints detected at different variance levels and Table III tabulates the average rate of change (absolute) in number of keypoints detected; the values in Table III essentially indicate the stability of the detectors under different image transformations; smaller values (darker cells) denotes lower sensitivity. In most cases FAST, AGAST, and ORB detected a much higher number of keypoints when a lower amount of variance was present, but at the same time, as more variance was introduced the number of keypoints detected was significantly reduced. Although this reduction was nearly proportionate to other detectors (Table III) under illumination variance, for variance in blur and jpeg-compression the differences were more significant.

For variance in jpeg compression, AKAZE, KAZE, Harris Laplace, and Star detector had almost zero change in detection rate. Even though all the detectors picked up pixelated edges and corners as keypoints, BRISK in particular was more prone to this error and as a result, detected more keypoints at the highest compression level.

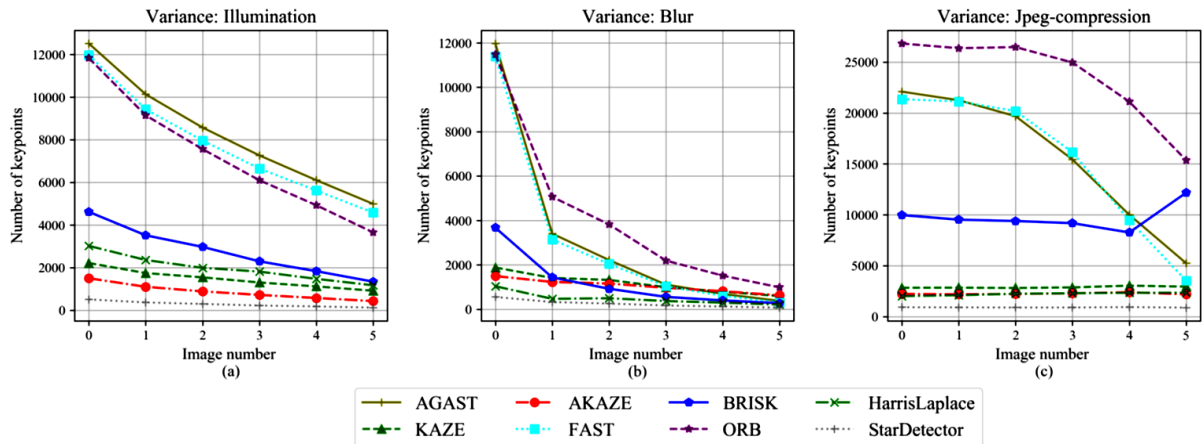

Figure 1. Detected keypoints over different levels of variance

### A. Number of Detected Keypoints

We start our evaluation by examining how photometric changes affect the detection capability of the detectors at

TABLE I. MEMORY SIZE OF THE DESCRIPTORS

| Descriptor | Size (bytes) |
|---|---|
| LUCID | 27 |
| LATCH | 32 |
| ORB | 32 |
| BoostDesc | 32 |
| BRIEF | 32 |
| AKAZE | 61 |
| FREAK | 64 |
| BRISK | 64 |

TABLE II. AVERAGE PROCESSING TIME TAKEN BY EACH DETECTOR

| Detector | Processing Time |
|---|---|
| FAST | 0.002 |
| AGAST | 0.007 |
| ORB | 0.01 |
| StarDetector | 0.02 |
| AKAZE | 0.07 |
| BRISK | 0.2 |
| KAZE | 0.33 |
| HarrisLaplace | 0.68 |

TABLE III. DETECTOR SENSITIVITY UNDER DIFFERENT LEVELS OF VARIANCE

| Variance / Detector | Illumination | Blur | Jpeg-compression |
|---|---|---|---|
| AGAST | 0.17 | 0.48 | 0.23 |
| KAZE | 0.16 | 0.2 | 0.025 |
| AKAZE | 0.22 | 0.16 | 0.039 |
| FAST | 0.17 | 0.49 | 0.26 |
| BRISK | 0.22 | 0.39 | 0.13 |
| ORB | 0.21 | 0.38 | 0.1 |
| HarrisLaplace | 0.17 | 0.27 | 0.032 |
| StarDetector | 0.24 | 0.32 | 0.027 |

### B. Repeatability Ratio

Next, we measured the detector performance in terms of repeatability ratio within the overlapping region **S**. The overlapping region is the part of the image in both **T** and **Q** that contains the same part of a scene. In other words, all the points in **T** that have corresponding points in **Q** form the overlapping region **S** (1).

$$S = Q \cap TH \quad (1)$$

**T_S** and **Q_S** denotes the part of the train image **T**, and query image **Q** containing the overlap region **S**. **K_TS**, **K_QS** denotes significant margin. Under different jpeg-compression levels, again KAZE and AKAZE produced better results. In the case of perspective, and zoom and rotation variance, ORB consistently outperformed other methods.

These results indicate that KAZE and AKAZE detectors are relatively more resilient to photometric changes while ORB is more resilient to geometric changes.

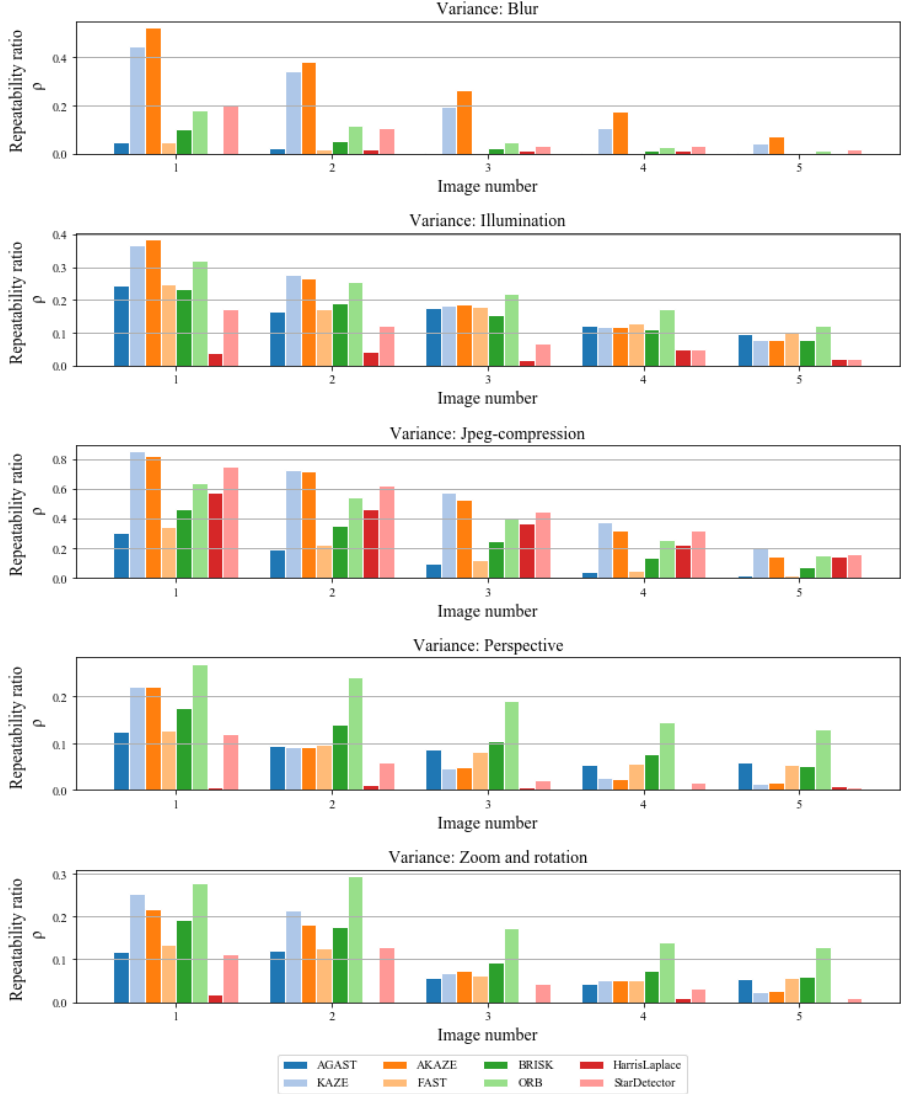

Figure 2. Repeatability ratio of the detectors under different levels of variances

the set of extracted keypoints from T_S and Q_S respectively. Using the label (homography matrix **H**) we computed the corresponding locations of keypoints in **K_TS** in **Q** and compared with **K_QS** to find the matching keypoints. The repeatability ratio $\rho$ (2) is the ratio of the total number of matched keypoints and the total number of keypoints detected in T_S.

$$\rho = \frac{\#(K_{TS} \cap K_{QS})}{\#K_{TS}} \quad (2)$$

Fig. 2 shows that KAZE and AKAZE had better repeatability ratios for lower variance in illumination but for higher variance ORB performed better. For blurred images, KAZE and AKAZE were the winners by a

*C. Processing Time*

Instead of computing execution time per keypoint, we measured the average execution time taken for each of the methods to process the entire image in the dataset. This information is more relevant from a development standpoint. For instance, a certain detecting algorithm having lower execution time per detected keypoint may generate a much higher number of keypoints, resulting in higher processing time for an entire image. The processing time was measured on a machine equipped with Intel®Core™i7-8750H Processor (2.20GHz, up to 4.10GHz) with 16-gigabyte memory.

Table II lists the average processing time for each of the detector methods in ascending order. We measured the processing time for feature extraction of each descriptor and detector combination under different photometric changes. Since the number of detected keypoints decreases at different rates for different detectors under photometric changes (as seen in section IV-A), the processing time to generate the descriptors would be different as well. Table IV shows the processing time for each descriptor when paired with all the other detectors. BRIEF, LUCID, and ORB were consistently the faster solutions under all the changes and BRISK was the slowest in most cases.

TABLE IV. AVERAGE EXECUTION TIME FOR DIFFERENT DETECTOR-DESCRIPTOR COMBINATIONS UNDER VARIANCES IN (A) BLUR, (B) ILLUMINATION, AND (C) JPEG-COMPRESSION. ROWS ARE DESCRIPTORS, COLUMNS ARE DETECTORS

|    | AG    | KZ    | AZ    | FT    | BK    | OB    | HL    | SD    |
|----|-------|-------|-------|-------|-------|-------|-------|-------|
| LT | 0.13  | 0.053 | 0.048 | 0.12  | 0.05  | 0.17  | 0.019 | 0.011 |
| LD | 0.004 | 0.003 | 0.003 | 0.003 | 0.003 | 0.004 | 0.002 | 0.002 |
| FK | 0.052 | 0.033 | 0.033 | 0.044 | 0.032 | 0.044 | 0.028 | 0.026 |
| BD | 0.13  | 0.049 | 0.045 | 0.13  | 0.054 | 0.19  | 0.021 | 0.012 |
| AZ | -     | 0.094 | 0.097 | -     | -     | -     | -     | -     |
| BF | 0.012 | 0.006 | 0.006 | 0.012 | 0.006 | 0.014 | 0.004 | 0.003 |
| BK | 0.19  | 0.15  | 0.15  | 0.17  | 0.15  | 0.18  | 0.14  | 0.14  |
| OB | 0.007 | 0.009 | 0.008 | 0.007 | 0.009 | 0.014 | 0.009 | 0.003 |

(A) Variance: Blur

|    | AG    | KZ    | AZ    | FT    | BK    | OB    | HL    | SD    |
|----|-------|-------|-------|-------|-------|-------|-------|-------|
| LT | 0.27  | 0.055 | 0.037 | 0.25  | 0.12  | 0.34  | 0.074 | 0.012 |
| LD | 0.005 | 0.002 | 0.002 | 0.005 | 0.003 | 0.005 | 0.002 | 0.002 |
| FK | 0.076 | 0.033 | 0.03  | 0.072 | 0.041 | 0.058 | 0.035 | 0.026 |
| BD | 0.33  | 0.061 | 0.035 | 0.32  | 0.12  | 0.32  | 0.079 | 0.013 |
| AZ | -     | 0.068 | 0.067 | -     | -     | -     | -     | -     |
| BF | 0.019 | 0.005 | 0.004 | 0.02  | 0.01  | 0.026 | 0.008 | 0.002 |
| BK | 0.2   | 0.14  | 0.14  | 0.2   | 0.17  | 0.21  | 0.15  | 0.14  |
| OB | 0.009 | 0.007 | 0.005 | 0.008 | 0.01  | 0.018 | 0.009 | 0.002 |

(B) Variance: Illumination

|    | AG    | KZ    | AZ    | FT    | BK    | OB    | HL    | SD    |
|----|-------|-------|-------|-------|-------|-------|-------|-------|
| LT | 0.56  | 0.11  | 0.093 | 0.54  | 0.38  | 0.96  | 0.086 | 0.037 |
| LD | 0.008 | 0.003 | 0.002 | 0.008 | 0.005 | 0.011 | 0.002 | 0.002 |
| FK | 0.11  | 0.039 | 0.037 | 0.11  | 0.078 | 0.18  | 0.036 | 0.029 |
| BD | 0.67  | 0.12  | 0.096 | 0.67  | 0.41  | 1     | 0.091 | 0.035 |
| AZ | -     | 0.067 | 0.07  | -     | -     | -     | -     | -     |
| BF | 0.037 | 0.008 | 0.008 | 0.038 | 0.027 | 0.063 | 0.008 | 0.004 |
| BK | 0.25  | 0.16  | 0.15  | 0.25  | 0.21  | 0.34  | 0.16  | 0.14  |
| OB | 0.017 | 0.008 | 0.007 | 0.016 | 0.016 | 0.031 | 0.008 | 0.003 |

(C) Variance: Jpeg-compression

### D. Accuracy, Precision, Recall

Finally, we analyzed the matching performance of all the possible detector-descriptor combinations. To match the keypoints we used K-nearest-neighbor and then further pruned the result using the distant ratio test [37]. We used the standard distant ratio of 0.7 for the test. We then categorized the keypoints in $\mathbf{K_{TS}}$ as unmatched and matched. Afterward, we used the homography model and an inlier distance threshold of 2.5 to determine good matches amongst the matched group of points; these keypoints were then classified as true positive and the rest as false positive. For the unmatched group, we executed the same procedure; keypoints that had a corresponding point on the image **Q** were then classified as false negative and the rest true negative. These results can vary depending on the initial matching technique, the values of the distance ratio, and the inlier distance threshold. We computed three metrics: *Accuracy*, *Precision*, and *Recall* (also known as inlier-ratio) using (3).

$$\left.\begin{array}{l} Accuracy = \dfrac{TP + TN}{TP + TN + FP + FN} \\ Precision = \dfrac{TP}{TP + FP} \\ Recall = \dfrac{TP}{TP + FN} \end{array}\right\} \quad (3)$$

$TP(True\ Positive), FP(False\ Positive),$
$TN(True\ Negative), FN(False\ Negative)$

The results for each of the detector-descriptor combinations are assembled in Table V. In almost all the cases AKAZE detector-descriptor performed better than other combinations. None of the descriptors performed particularly well under geometric changes, BRIEF and LATCH in particular had zero inliers. LUCID's performance was significantly poorer compare to other descriptors. In most cases, the descriptors showed better performance when paired with the KAZE or AKAZE detectors. Furthermore, AKAZE had better matching performance than all other descriptor methods.

## V. CONCLUSION

In this paper, we have presented a comparative evaluation of eight detectors and eight descriptors in the presence of different geometric and photometric transformations. The aim was to compare these methods in order to give an insight into which combinations may yield optimal performance in terms of speed and accuracy.

Amongst the detectors, FAST, AGAST, and ORB detected more keypoints on average and were significantly faster. This can be due to the fact that both AGAST and ORB detectors are variants of FAST. However, in terms of repeatability rate, KAZE and AKAZE showed to perform better under photometric changes, while ORB was more robust under geometric transformations. In general, descriptors performed better when paired with the KAZE and AKAZE detectors. As for the descriptors, BRIEF, LUCID, and ORB were faster than other methods while AKAZE consistently produced better results at matching keypoints.

### ACKNOWLEDGMENT

This work has been supported in part by the Office of Naval Research award N00014-16-1-2312 and US Army Research Laboratory (ARO) award W911NF-20-2-0084.

TABLE V.  AVERAGE ACCURACY, PRECISION, AND RECALL FOR DIFFERENT VARIANCES. ROWS ARE DESCRIPTORS, COLUMNS ARE DETECTORS

### Blur

| Accuracy | LT | LD | FK | BD | AZ | BF | BK | OB |
|---|---|---|---|---|---|---|---|---|
| AG | 0.05 | 0.15 | 0.01 | 0.01 | - | 0.07 | 0.01 | 0.04 |
| KZ | 0.17 | 0.15 | 0.12 | 0.07 | 0.4 | 0.24 | 0.11 | 0.19 |
| AZ | 0.13 | 0.13 | 0.12 | 0.06 | 0.41 | 0.26 | 0.12 | 0.11 |
| FT | 0.05 | 0.15 | 0.01 | 0.01 | - | 0.07 | 0.01 | 0.04 |
| BK | 0.03 | 0.13 | 0.05 | 0.02 | - | 0.08 | 0.07 | 0.04 |
| OB | 0.03 | 0.08 | 0.11 | 0.03 | - | 0.07 | 0.16 | 0.05 |
| HL | 0.05 | 0.21 | 0.03 | 0.02 | - | 0.09 | 0.03 | 0.05 |
| SD | 0.17 | 0.26 | 0.09 | 0.07 | - | 0.2 | 0.1 | 0.12 |

| Precision | LT | LD | FK | BD | AZ | BF | BK | OB |
|---|---|---|---|---|---|---|---|---|
| AG | 0.09 | 0.01 | 0.02 | 0.02 | - | 0.11 | 0.02 | 0.07 |
| KZ | 0.32 | 0.06 | 0.24 | 0.14 | 0.54 | 0.42 | 0.22 | 0.35 |
| AZ | 0.27 | 0.08 | 0.25 | 0.15 | 0.62 | 0.45 | 0.25 | 0.24 |
| FT | 0.1 | 0.01 | 0.02 | 0.02 | - | 0.12 | 0.02 | 0.07 |
| BK | 0.06 | 0.02 | 0.11 | 0.04 | - | 0.15 | 0.13 | 0.08 |
| OB | 0.07 | 0.01 | 0.19 | 0.04 | - | 0.14 | 0.27 | 0.11 |
| HL | 0.06 | 0.03 | 0.04 | 0.01 | - | 0.1 | 0.03 | 0.06 |
| SD | 0.31 | 0.14 | 0.19 | 0.15 | - | 0.35 | 0.22 | 0.23 |

| Recall \ Inlier-Ratio | LT | LD | FK | BD | AZ | BF | BK | OB |
|---|---|---|---|---|---|---|---|---|
| AG | 0.06 | 0.01 | 0.01 | 0.01 | - | 0.08 | 0.01 | 0.05 |
| KZ | 0.22 | 0.04 | 0.15 | 0.1 | 0.49 | 0.3 | 0.16 | 0.24 |
| AZ | 0.17 | 0.04 | 0.16 | 0.09 | 0.51 | 0.32 | 0.17 | 0.15 |
| FT | 0.07 | 0.01 | 0.01 | 0.01 | - | 0.08 | 0.01 | 0.05 |
| BK | 0.04 | 0.01 | 0.07 | 0.02 | - | 0.1 | 0.09 | 0.05 |
| OB | 0.05 | 0.01 | 0.14 | 0.03 | - | 0.1 | 0.19 | 0.08 |
| HL | 0.04 | 0.02 | 0.02 | 0.01 | - | 0.07 | 0.02 | 0.04 |
| SD | 0.21 | 0.08 | 0.12 | 0.09 | - | 0.24 | 0.13 | 0.15 |

### Illumination

| Accuracy | LT | LD | FK | BD | AZ | BF | BK | OB |
|---|---|---|---|---|---|---|---|---|
| AG | 0.19 | 0.08 | 0.06 | 0.07 | - | 0.2 | 0.12 | 0.17 |
| KZ | 0.17 | 0.08 | 0.1 | 0.13 | 0.29 | 0.21 | 0.18 | 0.19 |
| AZ | 0.13 | 0.1 | 0.11 | 0.11 | 0.27 | 0.23 | 0.2 | 0.13 |
| FT | 0.19 | 0.08 | 0.07 | 0.08 | - | 0.21 | 0.13 | 0.18 |
| BK | 0.07 | 0.08 | 0.09 | 0.03 | - | 0.2 | 0.14 | 0.07 |
| OB | 0.07 | 0.05 | 0.15 | 0.04 | - | 0.15 | 0.19 | 0.08 |
| HL | 0.07 | 0.17 | 0.02 | 0.02 | - | 0.09 | 0.02 | 0.05 |
| SD | 0.2 | 0.18 | 0.11 | 0.11 | - | 0.21 | 0.14 | 0.17 |

| Precision | LT | LD | FK | BD | AZ | BF | BK | OB |
|---|---|---|---|---|---|---|---|---|
| AG | 0.36 | 0 | 0.15 | 0.14 | - | 0.37 | 0.26 | 0.34 |
| KZ | 0.36 | 0 | 0.23 | 0.27 | 0.48 | 0.4 | 0.35 | 0.39 |
| AZ | 0.28 | 0 | 0.24 | 0.25 | 0.48 | 0.44 | 0.39 | 0.29 |
| FT | 0.37 | 0 | 0.15 | 0.15 | - | 0.39 | 0.27 | 0.36 |
| BK | 0.17 | 0 | 0.2 | 0.06 | - | 0.38 | 0.3 | 0.18 |
| OB | 0.18 | 0 | 0.31 | 0.07 | - | 0.33 | 0.36 | 0.2 |
| HL | 0.13 | 0 | 0.03 | 0.02 | - | 0.14 | 0.03 | 0.07 |
| SD | 0.41 | 0 | 0.24 | 0.26 | - | 0.42 | 0.31 | 0.37 |

| Recall \ Inlier-Ratio | LT | LD | FK | BD | AZ | BF | BK | OB |
|---|---|---|---|---|---|---|---|---|
| AG | 0.25 | 0 | 0.09 | 0.09 | - | 0.26 | 0.17 | 0.23 |
| KZ | 0.23 | 0 | 0.13 | 0.17 | 0.36 | 0.26 | 0.23 | 0.24 |
| AZ | 0.17 | 0 | 0.14 | 0.15 | 0.34 | 0.29 | 0.26 | 0.17 |
| FT | 0.26 | 0 | 0.1 | 0.1 | - | 0.27 | 0.18 | 0.25 |
| BK | 0.1 | 0 | 0.12 | 0.03 | - | 0.25 | 0.18 | 0.1 |
| OB | 0.1 | 0 | 0.16 | 0.04 | - | 0.2 | 0.23 | 0.11 |
| HL | 0.1 | 0 | 0.02 | 0.01 | - | 0.11 | 0.02 | 0.05 |
| SD | 0.25 | 0 | 0.13 | 0.13 | - | 0.25 | 0.18 | 0.22 |

### Jpeg-compression

| Accuracy | LT | LD | FK | BD | AZ | BF | BK | OB |
|---|---|---|---|---|---|---|---|---|
| AG | 0.15 | 0.04 | 0.06 | 0.04 | - | 0.17 | 0.08 | 0.13 |
| KZ | 0.37 | 0.1 | 0.28 | 0.16 | 0.74 | 0.42 | 0.33 | 0.39 |
| AZ | 0.3 | 0.12 | 0.27 | 0.14 | 0.56 | 0.48 | 0.36 | 0.3 |
| FT | 0.18 | 0.05 | 0.07 | 0.05 | - | 0.19 | 0.1 | 0.15 |
| BK | 0.1 | 0.06 | 0.15 | 0.03 | - | 0.23 | 0.2 | 0.12 |
| OB | 0.14 | 0.04 | 0.34 | 0.04 | - | 0.18 | 0.38 | 0.18 |
| HL | 0.23 | 0.12 | 0.16 | 0.09 | - | 0.26 | 0.19 | 0.24 |
| SD | 0.48 | 0.17 | 0.35 | 0.19 | - | 0.51 | 0.38 | 0.41 |

| Precision | LT | LD | FK | BD | AZ | BF | BK | OB |
|---|---|---|---|---|---|---|---|---|
| AG | 0.36 | 0.02 | 0.18 | 0.13 | - | 0.4 | 0.22 | 0.32 |
| KZ | 0.68 | 0.1 | 0.61 | 0.41 | 0.92 | 0.76 | 0.58 | 0.71 |
| AZ | 0.59 | 0.2 | 0.56 | 0.38 | 0.84 | 0.78 | 0.61 | 0.6 |
| FT | 0.39 | 0.03 | 0.22 | 0.15 | - | 0.44 | 0.25 | 0.36 |
| BK | 0.27 | 0.07 | 0.39 | 0.09 | - | 0.48 | 0.42 | 0.31 |
| OB | 0.33 | 0.05 | 0.74 | 0.14 | - | 0.43 | 0.67 | 0.4 |
| HL | 0.48 | 0.13 | 0.41 | 0.25 | - | 0.53 | 0.44 | 0.51 |
| SD | 0.82 | 0.32 | 0.71 | 0.53 | - | 0.86 | 0.71 | 0.75 |

| Recall \ Inlier-Ratio | LT | LD | FK | BD | AZ | BF | BK | OB |
|---|---|---|---|---|---|---|---|---|
| AG | 0.18 | 0.01 | 0.07 | 0.05 | - | 0.19 | 0.09 | 0.14 |
| KZ | 0.39 | 0.05 | 0.3 | 0.18 | 0.76 | 0.45 | 0.36 | 0.41 |
| AZ | 0.33 | 0.05 | 0.3 | 0.15 | 0.59 | 0.51 | 0.38 | 0.33 |
| FT | 0.2 | 0.01 | 0.09 | 0.06 | - | 0.22 | 0.12 | 0.17 |
| BK | 0.13 | 0.03 | 0.18 | 0.03 | - | 0.27 | 0.24 | 0.15 |
| OB | 0.16 | 0.02 | 0.35 | 0.05 | - | 0.21 | 0.41 | 0.21 |
| HL | 0.26 | 0.05 | 0.18 | 0.1 | - | 0.29 | 0.22 | 0.27 |
| SD | 0.5 | 0.08 | 0.36 | 0.2 | - | 0.52 | 0.39 | 0.43 |

### Perspective

| Accuracy | LT | LD | FK | BD | AZ | BF | BK | OB |
|---|---|---|---|---|---|---|---|---|
| AG | 0 | 0.08 | 0.01 | 0 | - | 0.01 | 0.01 | 0.02 |
| KZ | 0 | 0.1 | 0.02 | 0 | 0 | 0.01 | 0.02 | 0.01 |
| AZ | 0 | 0.11 | 0.02 | 0.02 | 0.03 | 0.03 | 0.03 | 0.02 |
| FT | 0 | 0.08 | 0.01 | 0 | - | 0.01 | 0.01 | 0.02 |
| BK | 0 | 0.08 | 0.03 | 0.01 | - | 0 | 0.03 | 0.01 |
| OB | 0 | 0.05 | 0.02 | 0 | - | 0 | 0.02 | 0.02 |
| HL | 0 | 0.13 | 0 | 0 | - | 0 | 0.01 | 0.01 |
| SD | 0 | 0.19 | 0.02 | 0 | - | 0.01 | 0.02 | 0.01 |

| Precision | LT | LD | FK | BD | AZ | BF | BK | OB |
|---|---|---|---|---|---|---|---|---|
| AG | 0 | 0.12 | 0.25 | 0.05 | - | 0.34 | 0.19 | 0.25 |
| KZ | 0 | 0.2 | 0.4 | 0.07 | 0.02 | 0.21 | 0.37 | 0.41 |
| AZ | 0.12 | 0.84 | 0.37 | 0.64 | 0.45 | 0.44 | 0.49 | 0.45 |
| FT | 0 | 0.11 | 0.47 | 0.01 | - | 0.33 | 0.21 | 0.27 |
| BK | 0.28 | 0.35 | 0.79 | 0.42 | - | 0.21 | 0.68 | 0.65 |
| OB | 0.27 | 0.69 | 0.6 | 0.26 | - | 0.21 | 0.45 | 0.43 |
| HL | 0 | 0.1 | 0.21 | 0 | - | 0 | 0 | 0 |
| SD | 0 | 0.76 | 0.36 | 0.22 | - | 0.42 | 0.57 | 0.26 |

| Recall \ Inlier-Ratio | LT | LD | FK | BD | AZ | BF | BK | OB |
|---|---|---|---|---|---|---|---|---|
| AG | 0 | 0 | 0.01 | 0 | - | 0 | 0.01 | 0 |
| KZ | 0 | 0 | 0.01 | 0 | 0 | 0 | 0.02 | 0 |
| AZ | 0 | 0.01 | 0.02 | 0.02 | 0.03 | 0 | 0.02 | 0.01 |
| FT | 0 | 0 | 0.01 | 0 | - | 0 | 0.01 | 0 |
| BK | 0 | 0 | 0.03 | 0.01 | - | 0 | 0.03 | 0.01 |
| OB | 0 | 0 | 0.02 | 0 | - | 0 | 0.02 | 0.01 |
| HL | 0 | 0 | 0 | 0 | - | 0 | 0 | 0 |
| SD | 0 | 0.01 | 0.01 | 0 | - | 0 | 0.02 | 0 |

### Scale and Rotation

| Accuracy | LT | LD | FK | BD | AZ | BF | BK | OB |
|---|---|---|---|---|---|---|---|---|
| AG | 0 | 0.1 | 0 | 0 | - | 0.01 | 0 | 0 |
| KZ | 0 | 0.12 | 0.01 | 0 | 0 | 0 | 0.02 | 0 |
| AZ | 0 | 0.13 | 0.01 | 0.01 | 0.07 | 0.01 | 0.02 | 0.01 |
| FT | 0 | 0.1 | 0 | 0 | - | 0.01 | 0 | 0 |
| BK | 0 | 0.1 | 0.02 | 0.01 | - | 0 | 0.04 | 0.01 |
| OB | 0 | 0.08 | 0.03 | 0 | - | 0 | 0.04 | 0.01 |
| HL | 0 | 0.16 | 0 | 0 | - | 0 | 0 | 0 |
| SD | 0 | 0.18 | 0.02 | 0 | - | 0.01 | 0.02 | 0.01 |

| Precision | LT | LD | FK | BD | AZ | BF | BK | OB |
|---|---|---|---|---|---|---|---|---|
| AG | 0 | 0.6 | 0.01 | 0 | - | 0 | 0.02 | 0 |
| KZ | 0 | 0.41 | 0.7 | 0.2 | 0.2 | 0.01 | 0.53 | 0 |
| AZ | 0.02 | 0.23 | 0.52 | 0.75 | 0.88 | 0.01 | 0.78 | 0.24 |
| FT | 0 | 0.4 | 0.02 | 0 | - | 0 | 0.23 | 0 |
| BK | 0 | 0.6 | 0.76 | 0.44 | - | 0.2 | 0.79 | 0.25 |
| OB | 0 | 0.6 | 0.69 | 0.41 | - | 0.2 | 0.83 | 0.71 |
| HL | 0 | 0.4 | 0.2 | 0 | - | 0 | 0.2 | 0.2 |
| SD | 0 | 0.46 | 0.79 | 0.2 | - | 0.01 | 0.62 | 0 |

| Recall \ Inlier-Ratio | LT | LD | FK | BD | AZ | BF | BK | OB |
|---|---|---|---|---|---|---|---|---|
| AG | 0 | 0 | 0 | 0 | - | 0 | 0 | 0 |
| KZ | 0 | 0 | 0.01 | 0 | 0 | 0 | 0.02 | 0 |
| AZ | 0 | 0 | 0.01 | 0.01 | 0.07 | 0 | 0.02 | 0.01 |
| FT | 0 | 0 | 0 | 0 | - | 0 | 0 | 0 |
| BK | 0 | 0 | 0.02 | 0 | - | 0 | 0.04 | 0.01 |
| OB | 0 | 0 | 0.02 | 0 | - | 0 | 0.03 | 0.01 |
| HL | 0 | 0 | 0 | 0 | - | 0 | 0 | 0 |
| SD | 0 | 0 | 0.02 | 0 | - | 0 | 0.02 | 0 |

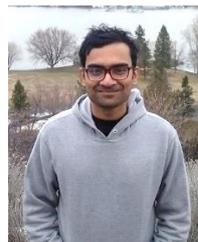

**Shuvo Kumar Paul** received his M.S. in 2020 in computer science and engineering from University of Nevada, Reno. He completed his B.S. from North South University, Bangladesh. Before joining UNR, he worked as a research associate at the AGENCY lab (previously CVCR). His research interests include machine learning, computer vision, computational linguistics, and robotics

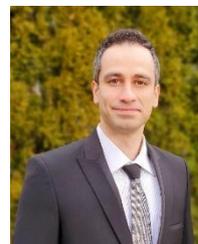

**Pourya Hoseini** received his Ph.D. in 2020 and M.S. in 2017, both in computer science and engineering from University of Nevada, Reno. He also received a M.S. degree in 2011 and a B.S. in 2008 in electrical engineering from Urmia University and Azad University, Iran, respectively. His research interests are machine learning, computer vision, and evolutionary computing.

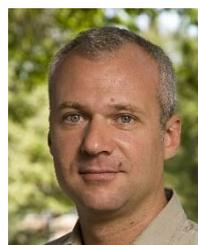

**Mircea Nicolescu** is a Professor of Computer Science and Engineering at the University of Nevada, Reno and co-director of the UNR Computer Vision Laboratory. He received a PhD degree from the University of Southern California in 2003, a MS degree from USC in 1999, and a BS degree from the Polytechnic University Bucharest, Romania in 1995, all in Computer Science. His research interests include visual motion analysis, perceptual organization, vision-based surveillance, and activity recognition. Dr. Nicolescu's research has been funded by the Department of Homeland Security, the Office of Naval Research, the National Science Foundation and NASA. He is a member of the IEEE Computer Society.

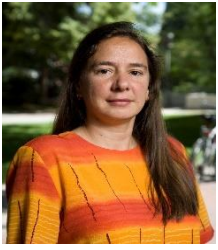
**Dr. Monica Nicolescu** is a Professor with the Computer Science and Engineering Department at the University of Nevada, Reno and is the Director of the UNR Robotics Research Lab. Dr. Nicolescu earned her PhD degree in Computer Science from the University of Southern California (2003) at the Center for Robotics and Embedded Systems. She obtained her MS degree in Computer Science from USC (1999) and a BS in Computer Science at the Polytechnic University Bucharest (Romania, 1995). Her research interests are in the areas of human-robot interaction, robot control, learning, and multi-robot systems. Dr. Nicolescu's research has been supported by the National Science Foundation, the Office of Naval Research, the Army Research Laboratory, the Department of Energy and Nevada Nanotech Systems. In 2006 she was a recipient of the NSF Early Career Development Award (CAREER) Award for her work on robot learning by demonstration